\documentclass[runningheads]{llncs}
\usepackage[T1]{fontenc}
\usepackage{multirow}
\usepackage{graphicx}
\usepackage{hyperref}
\usepackage{color}

\usepackage{amsmath}
\usepackage{amssymb}
\usepackage{mathtools}
\usepackage[toc]{appendix}

\usepackage{bbding} 
\begin{document}
\title{When Can You Trust Your Explanations? \\A Robustness Analysis on Feature Importances}
\titlerunning{When Can You Trust Your Explanations?}
%
\author{Ilaria Vascotto\inst{1}\Envelope\orcidID{0009-0004-3961-1674} \and
Alex Rodriguez\inst{1,2}\orcidID{0000-0002-0213-6695} \and Alessandro Bonaita\inst{3} \and
Luca Bortolussi\inst{1}\orcidID{0000-0001-8874-4001}}
\authorrunning{I. Vascotto et al.}
%
\institute{Department of Mathematics, Informatics and Geosciences, \newline University of Trieste, Trieste, Italy \newline
\email{ilaria.vascotto@phd.units.it}
\newline \email{\{alejandro.rodriguezgarcia, lbortolussi\}@units.it}
\and
The Abdus Salam International Center for Theoretical Physics, Trieste, Italy \and 
Assicurazioni Generali Spa, Milan, Italy 
\newline \email{alessandro.bonaita@generali.com}
}
\maketitle              
\begin{abstract}  
Recent legislative regulations have underlined the need for accountable and transparent artificial intelligence systems and have contributed to a growing interest in the Explainable Artificial Intelligence (XAI) field. Nonetheless, the lack of standardized criteria to validate explanation methodologies remains a major obstacle to developing trustworthy systems. We address a crucial yet often overlooked aspect of XAI, the robustness of explanations, which plays a central role in ensuring trust in both the system and the provided explanation. To this end, we propose a novel approach to analyse the robustness of neural network explanations to non-adversarial perturbations, leveraging the manifold hypothesis to produce new perturbed datapoints that resemble the observed data distribution. We additionally present an ensemble method to aggregate various explanations, showing how merging explanations can be beneficial for both understanding the model’s decision and evaluating  the robustness. The aim of our work is to provide practitioners with a framework for evaluating the trustworthiness of model explanations. Experimental results on feature importances derived from neural networks applied to tabular datasets highlight the importance of robust explanations in practical applications.

\end{abstract}
\section{Introduction} \label{introduction}
The popularity of neural networks and their application to high-risk scenarios has recently raised questions on their accountability and trustworthiness. The rapid expansion of the field of Artificial Intelligence (AI) has stimulated legislative discussions and a series of novel regulations and guidelines has been proposed. In the European Union, the \textit{AI Act} \cite{aiact} and parts of the \textit{General Data Protection Regulation} (GDPR) \cite{gdpr} have stressed the need for a fairer and more transparent approach to artificial intelligence. Similarly, the United States of America have proposed the new \textit{Blueprint for an AI bill of rights} \cite{blueprint} that strives for a fair development and deployment of AI systems.

Modern AI systems are increasingly complex due to the elevated number of parameters involved to solve challenging tasks and are often referred to as \textit{black boxes} given the opaque nature of their predictions. Transparency is, instead, a fundamental property that AI systems should guarantee, aiming to provide detailed descriptions of the model reasoning, even in natural language. Imagine that a model is being used in healthcare for patient diagnosis. 
If doctors can understand how it reached a given prediction, they gain a valuable tool for evaluating the correctness of such diagnosis and grow confidence in the AI system, even if they don't fully comprehend the technical aspects of the \textit{black box} model.

The field of Explainable Artificial Intelligence (XAI) has proposed a variety of approaches to \textit{open the black box} and provide explanations addressing the model inner reasoning, for example in the form of feature importances. While not explicitly referred to from a legislative standpoint, XAI can act as a powerful tool in enhancing transparency of AI-based systems. Understanding the reasoning behind a model decision is an asset from a technical standpoint, allowing experts to validate the predictions and detect possible biases before a model is deployed. Additionally, end-users may benefit from explanations as the \textit{right to explanation} cited in the GDPR explicitly requires the user to receive an explanation when the decision is entirely subject to an automated decision system. 

Despite their usefulness, the lack of standardized criteria to validate explainability approaches is still a major obstacle towards transparent and trustworthy systems. Although it is a critical aspect, the robustness of explanations remains an often underexplored facet of the development of explanation approaches. Robustness can be defined as an explainer’s ability to provide consistent explanations for similar inputs. It can be evaluated through both non-adversarial perturbations, showing intrinsic weaknesses even under small changes, and adversarial attacks, implying a malicious nature of manipulating explanations. 

Another complex characteristic of the XAI field is the disagreement problem \cite{krishna-lakkaraju}, which occurs in scenarios where multiple explanation methods applied to the same datapoint return contrasting results. The debate over which explanation to choose (or \textit{trust}) is still open, as explanation disagreement posits practical impediments to trustable AI systems.

Our contribution explores the following points: 
\begin{itemize}
    \item We propose a set of desirable properties that a robustness estimator should satisfy and show that our proposal, tailored for feature importance methods, satisfies them all.
    \item We address the disagreement problem on neural network explanations by proposing an ensemble of explanations focused on the ranking of the features. 
    \item We introduce a framework to test explanation robustness to non-adversarial perturbations and assess their trustworthiness in practical applications. 
    \item We propose a novel validation assessment of the robustness estimation, to tackle the lack of a ground truth.
\end{itemize}

We have tested our proposal on eight publicly-available tabular datasets and three neural network-specific feature importance methods. Our analysis demonstrates the need for evaluation tools to assess explanation robustness, supporting transparency and accountability in real-world applications.

\section{Related Work} \label{related-work}

LIME \cite{lime} and SHAP \cite{shap} are among the most widely used XAI techniques in real-world applications. LIME works by fitting an inherently transparent model (such as a linear model or a decision tree) around the datapoint which is being explained. A neighbourhood is constructed from a fixed data distribution, generating a set of points on which the model, which acts as a local explanation, is fitted.
SHAP makes use of Shapley values to explain predictions, measuring how the prediction changes when a feature is included or excluded in the feature set. It then averages these changes across all possible combinations of features, producing a vector of feature importances. 

Despite their wide use, both methods lack \textit{robustness} (or \textit{stability}), which in this context represents the ability of an explanation method to produce similar and consistent explanations when different conditions change. LIME is an unstable method by design, as the neighbourhood generation step yields different sets of datapoints at each call of the method. This implies that, at each time the method is applied to the same datapoint, a different model is fitted, resulting in explanations whose coefficients differ feature-wise by magnitude or even by sign. SHAP is instead susceptible to feature correlations, sampling variability and data distribution shifts due to the way Shapley values are approximated. The instability of these approaches was first proved in \cite{fooling_lime_shap}, where the authors showed how an adversarial model could easily be defined to mask biased classifiers though unbiased explanations. Their untrustworthiness, along with other model-agnostic additive methods, has also been investigated in \cite{gosieska}. Their theoretical assumptions on feature independence are hardly met in practice, rendering them unable to correctly detect feature interactions when present.

According to \cite{mishra}, explanation robustness can be tested along three directions: robustness to input perturbations, to model changes and to hyperparameter selection. The first one includes the scenarios where the input may be modified by random perturbations or by adversarial attacks to the explanations themselves. The second one refers to manipulations of the model, such as fine-tuning of the parameters with a modified loss function, and the last one considers the influence of technique-specific hyperparameters.  

An adversarial attack in the context of XAI is a perturbation of the input such that the model prediction is unchanged but the explanation marks different features as important or not \cite{baniecki}. An evaluation is provided in \cite{ghorbani}, in which attacks to interpretations provided by neural networks aim at maximising the change in the explanations (according to different change functions) under the constraint that the perturbation is small and the prediction is unchanged. \cite{dimanov} explores how model manipulation, in particular with the introduction of a penalty term in the loss function, can damage explanations. They call for a rigorous test of robustness to limit the effects of such  manipulations. In \cite{fairwashing} the authors note that it is possible, for any classifier $g$, to construct a classifier $\hat g$ that exhibits the same behaviour on seen data but presents biased explanations. They propose a version of the presented gradient-based explanations (\cite{lrp,ig}) robust to model manipulation by projecting the explanations on the tangent space of the data manifold.

While random perturbations and adversarial attacks to explanations in the context of images (as in \cite{ghorbani}) can be easily examined by human experts, as the changes are often evident even to the naked eye, it is not sufficient to rely on non-quantitative evaluations. Robustness analysis must be supported by an adequate robustness estimation and agreed-upon metrics must be defined. To this end, \cite{kantz} proposes a robustness score to evaluate explanation robustness when the data generation process is known, but this assumption makes it difficult to adapt to real-world datasets, where a ground truth is hardly available. In \cite{alvarezmelis} the authors propose a formalization of local robustness based on the estimate of the local Lipschitz continuity. They test their proposal on explanations applied to images and show that gradient-based approaches are much more robust than their perturbation counterparts (LIME and SHAP).

The work of \cite{Nauta_2023} presents an in-depth survey on the evaluation of XAI techniques and identifies different metrics that can be used to assess explanation robustness. Similarities can be computed with metrics such as: rank order correlations, top-$k$ intersections, rule matching and structural similarity indexes. \cite{rosenfeld} proposes similar metrics for evaluating stability, as the Jaccard similarity, additionally requiring that stability tests should be performed using perturbations that do not change the class label and that introduce small amounts of resampling noise to ensure the stability of the explanations.

Robust-by-construction approaches have emerged as an interesting area of research. For example, \cite{self-explaining-nn} proposes self-explaining neural networks, a class of models for which faithfulness and robustness are enforced by construction through a specific regularization, aided by a generalization of the above-mentioned local Lipschitz continuity. Similarly, ROPE \cite{rope} is a framework based on adversarial training that generates explanations which are robust to both changes in the input and in the data distribution. 

We aim at investigating the robustness of explanations to non adversarial perturbations on tabular datasets, presenting a robustness metric that addresses limitations of the previously mentioned methods (recalled within brackets in this paragraph). In particular we define a metric that can be computed even in absence of a ground truth \cite{kantz}, considering practitioners needs (for example, deploying a model without requiring a retraining \cite{dimanov}) and keeping in mind the limitations derived from theoretical assumptions \cite{gosieska}. Importantly, our metric is bounded within the $[0,1]$ range, allowing for comparability among methods, datasets and used models \cite{alvarezmelis}.

Alongside robustness analysis, we seek to explore the efficacy of an ensemble approach on explanations. \cite{krishna-lakkaraju} presents an overview of the disagreement problem according to practitioners: while it is underlined that the problem is of non-negligible size, the authors do not propose solutions or good practices. In the context of adversarial attacks, \cite{rieger} showcases the efficacy of ensemble methods as defences on explanations. They test their aggregated explanation on image data and show that it is more resilient to attacks, when a composing explanation method or the model itself is being fooled. Aggregations can also be performed leveraging multiple explanations from the same method: \cite{bhatt} derives a more robust Shapley-value explanation by aggregating the explanations computed on a carefully crafted neighbourhood, minimizing explanation sensitivity. We will consider these characteristics when devising our ensemble and considering the robustness estimator.

\section{Background} \label{background}
This section introduces key terminology related to the XAI field and describes the techniques used in the experimental analysis.

\subsection{Terminology}
While there is not an agreed-upon taxonomy to classify XAI techniques, we follow the proposal of \cite{survey-adadi}, which identifies the following axes of interest:
\begin{itemize}
    \item Scope of the explanation: a local approach aims at explaining how a given individual prediction is made while a global one focuses on the model as a whole, analysing its overall reasoning.
    \item Model of interest: model-specific techniques are tailored to the structure of the model under investigation, while model-agnostic ones can be applied to any model.
    \item Transparency: intrinsically transparent models are interpretable by construction (and are also known as \textit{glass boxes}) while post-hoc techniques are applied after the model is fully trained.    
\end{itemize}

As we will be discussing results obtained from neural networks, we can further distinguish between perturbation-based and gradient-based approaches \cite{survey-nn}. Perturbation-based approaches are often model-agnostic and rely either on neighbourhood generation or combinatorial aspects, as in LIME \cite{lime} or SHAP \cite{shap}. Gradient-based methods, instead, are specific to neural networks and harness their inner structure, mainly taking advantage of the backpropagation mechanisms \cite{lrp,deeplift,ig}. 

The broader category of approaches we will be considering is that of feature attributions. Having an input $\mathbf{x}=(x_1, \dots x_m)$ with $m$ features, a feature attribution is a vector $\mathbf{a} = (a_1, \dots, a_m)$ of size $m$ where each entry represents the importance of the corresponding feature towards the model's prediction. According to the specific datatype of $\mathbf{x}$, the attributions may refer to individual variables (tabular data), words or bag of words (natural language), pixels or superpixels (images - in this case explanations are often called \textit{heatmaps}). A positive (negative) sign usually represents a positive (negative) contribution of the feature and its relevance in the prediction. In the following, we will use the terms \textit{feature importances} and \textit{attributions} interchangeably.

\subsection{Considered Techniques} \label{considered_techniques}
We focus on local post-hoc approaches specific to neural networks and that are applicable to nets trained on tabular datasets. We have not considered model-agnostic approaches such as LIME \cite{lime} and SHAP \cite{shap} due to their known instability \cite{alvarezmelis,gosieska,fooling_lime_shap}, as presented in Section \ref{related-work}. Model-specific approaches, such as Saliency and Input X Gradient \cite{survey-nn} are also known for being unstable \cite{alvarezmelis,ghorbani} and have therefore been excluded from the analysis.

Note that neural network explanations require the selection of the output neuron to be explained: this is more relevant in classification problems where one may want to investigate the features that contributed to any of the classes probability scores. In the following, the target neuron will be chosen as the one associated with the model's predictions, that is, the one with the largest output score. 

\subsubsection{DeepLIFT}
DeepLearning Important Features (DeepLIFT) \cite{deeplift} computes feature importances with respect to their difference from a given reference. In particular, the differences between the two outputs are explained in terms of the differences among the two inputs. Each neuron is analysed with respect to the difference between its activation and that of the reference input. DeepLIFT makes use of contribution scores and multipliers to backpropagate the difference in output through the network. It requires a single forward-backward pass through the net, making it efficient. The propagations are computed through appropriate chain rules, defined according to the neuron's type and its activations. 

\subsubsection{Integrated Gradients}
Integrated Gradients (IG) \cite{ig} satisfies the axioms of sensitivity and implementation invariance. It computes the integral of the gradients of a net $f$ along the straight-line path from a baseline $x'$ to the input $x$, considering a series of linearly separated instances along the path from the baseline to the point of interest. In practice, it takes advantage of an approximation of $s$ steps such that, for the $j$-th dimension, it holds:
\begin{equation}
\text{IG}_j^{approx} = \cfrac{x_j - x_j'}{s} \sum_{k=1}^{s} \cfrac{\partial f(x'+(x-x')\cdot k/s)}{\partial x_j}
\end{equation}
The authors of \cite{ig} found that $s\in(20,300)$ produced satisfactory approximations but the computation can nonetheless be expensive when the number of steps is large. 

\subsubsection{Layerwise Relevance Propagation}
Layerwise Relevance Propagation (LRP) \cite{lrp} is based on the backpropagation principle. It defines a series of rules to propagate the output score (or \textit{relevance}) $f(x)$ through the net's layers, according to the architecture at hand. The conservation property holds: with $R_j$ the relevance for neuron $j$, for each pair of layers $\sum _jR_j = \sum _k R_k$ and globally it holds that summing over all layers $\sum _i R_i = f(x)$. Two common propagation rules are the \texttt{epsilon} and the \texttt{gamma} rules:
\begin{equation}
\begin{split}
    \text{LRP-}\epsilon: R_j &= \sum_k \cfrac{a_j \cdot w_{jk}}{\epsilon + \sum_{0,j} a_j \cdot w_{jk}}R_k
    \\
    \text{LRP-}\gamma: R_j &= \sum_k \cfrac{a_j \cdot (w_{jk}+\gamma\cdot w_{jk}^+)}{\epsilon + \sum_{0,j} a_j \cdot (w_{jk} + \gamma \cdot w_{jk}^+)}R_k
\end{split}
\end{equation}

where $a_j$ is the activation of neuron $j$, $w_{jk}$ is the weight linking neuron $j$ to neuron $k$ in the following layer ($w_{jk}^+$ is a positive weight), $\sum _{0,j}$ is the sum over all lower-layer activations.

\paragraph{Missingness Property.} DeepLIFT, Integrated Gradients and LRP satisfy the missingness property: if $x_j = 0 \Rightarrow a_j = 0$. The features corresponding to the null entries in the feature vector will have null coefficients in the attribution vector, representing a lack of importance towards the prediction. This property is particularly relevant when dealing with categorical variables, preprocessed with one-hot encoding.

\section{Methodology} \label{methodology}
Our approach is applied to tabular datasets and classification problems. Let us introduce the following notation: let $\mathcal{D}=(\mathbf{X}, \mathbf{y})$ be a dataset with $N$ datapoints and $m$ features such that $(\mathbf{x}^i, y^i) = (x^{(i,1)}, \dots, x^{(i,j)}, \dots, x^{(i,m)}, y^i)$ with $y^i$ a class label. The dataset is split into a training, validation and test datasets, identified by $\mathcal{D}_{train}$, $\mathcal{D}_{valid}$ and $\mathcal{D}_{test}$ respectively. Let $f(\cdot)$ be a neural network trained on $\mathcal{D}_{train}$ and $t$ be a target class. Let $e$ be an explanation method (or explainer) and $e(\mathbf{x}^i) \coloneqq e(\mathbf{x}^i, f)$ the explanation of model $f$ prediction of point $\mathbf{x}^i$. More specifically, let the feature attribution vector of the $l$-th method be $\mathbf{a}_l^i = (a_l^{(i,1)}, \dots, a_l^{(i,m)})$.

\subsection{Robustness Estimator}
We define the robustness of an explanation as a measure of its variability when the input is modified. If we consider $\mathbf{x}$ the original datapoint, $\tilde{\mathbf{x}}$ a perturbation, $e$ an explanation method and $e(\mathbf{x})$ the corresponding explanation, then:
\begin{equation}
    \mathbf{x} \to \tilde{\mathbf{x}} , e(\mathbf{x}) \to e(\tilde{\mathbf{x}}) \Rightarrow r(\mathbf{x}, e) = g(\mathbf{x}, \tilde{\mathbf{x}}, e)
\end{equation}
that is, the robustness $r$ of $e(\mathbf{x})$ is a function of the chosen explanation method, the original datapoint and the perturbed one.

Further considering a constraint of the form $dist(\mathbf{x}, \tilde{\mathbf{x}}) < \epsilon$ with $\epsilon>0$ and $dist$ a distance metric (such as the euclidean distance), we can introduce the notion of local robustness. 
\begin{definition}
    An explanation method $e$ is locally robust if perturbing the input results in a similar explanation. If $\mathbf{x}\to \tilde{\mathbf{x}}$ with $dist(\mathbf{x}, \tilde{\mathbf{x}})<\epsilon \; (\epsilon > 0) \;\Rightarrow e(\mathbf{x}) \approx e(\tilde{\mathbf{x}})$.
\end{definition}

Given a robustness estimator $\hat{\mathcal{R}}(\mathbf{x}^i, \mathcal{N}^i) \coloneqq \hat{\mathcal{R}}(\mathbf{x}^i, \mathcal{N}^i, e, f)$ that measures the robustness of the explanation method $e$ applied to the neighbourhood $\mathcal{N}^i$ of the point $\mathbf{x}^i$ over the model $f$, we can define the following set of \textit{desiderata}.

\begin{property} \label{p1}
If $r(\mathbf{x}^i, \tilde{\mathbf{{x}}}^i) \coloneqq r(\mathbf{x}^i, \tilde{\mathbf{{x}}}^i, e, f)$ is the robustness of $e(\mathbf{x}^i)$ with respect to the perturbation $\tilde{\mathbf{x}}^i$, then the robustness $\mathcal{R} = \mathbb{E}[r]$ is estimated by:
\begin{equation}
\hat{\mathcal{R}}(\mathbf{x}^i, \mathcal{N}^i) = \cfrac{1}{\vert \mathcal{N}^i\vert} \sum\limits_{\tilde{\mathbf{x}}^i\in \mathcal{N}^i} r(\mathbf{x}^i,\tilde{\mathbf{x}}^i)
\end{equation}
where $\mathcal{N}^i = \{\tilde{\mathbf{x}}^i \vert \tilde{\mathbf{x}}^i = \mathbf{x}^i + \lambda \text{ with } dist(\mathbf{x}^i, \tilde{\mathbf{x}}^i)<\epsilon \; (\epsilon>0), \;  \lambda\in \mathbb{R}^m \}$.
\end{property}

Two points which are close to each other within the dataspace will produce explanations with comparable robustness scores.

\begin{property} \label{p2}
If $\hat{\mathcal{R}}$ is a local robustness estimator, then for two distinct points $\mathbf{x}^i, \mathbf{x}^j$ such that $dist(\mathbf{x}^i, \mathbf{x}^j)<\epsilon \; (\epsilon>0)$ it holds that:
\begin{equation}
    \exists \, \delta >0 \text{ s.t. }\vert \hat{\mathcal{R}}(\mathbf{x}^i, \mathcal{N}^i) - \hat{\mathcal{R}}(\mathbf{x}^j, \mathcal{N}^j)\vert < \delta 
\end{equation}
\end{property}

Robustness estimation is intrinsically linked to uncertainty in the estimates, due to errors in the explanations themselves and the lack of a ground truth for the robustness.

\begin{property} \label{p3}
$\hat{\mathcal{R}}$ is such that $\hat{\mathcal{R}} = \mathcal{R} + \theta_\epsilon$ where $\mathcal{R} =\mathbb{E}[r]$ is the true robustness and $\theta_\epsilon \neq 0$ is an error term.
\end{property}

By definition, the neighbourhood generation is highly influential in the robustness estimation process when non-adversarial perturbations are considered. On-manifold perturbations better reflect the true data distribution and the manifold learned by the model, therefore exhibit greater robustness scores than random perturbations which may be off-manifold. 

\begin{property} \label{p4}
If $\mathcal{N}^i$ is an on-manifold neighbourhood and $\bar{\mathcal{N}}^i$ an off-manifold one, then $\hat{\mathcal{R}}(\mathbf{x}^i, \mathcal{N}^i)> \hat{\mathcal{R}}(\mathbf{x}^i, \bar{\mathcal{N}}^i)$.
\end{property}

The robustness of an aggregation of explainers is bounded by the robustness of the individual components.

\begin{property} \label{p5}
If the explainer $\bar e$ is an aggregation of explainers, $\bar e = agg(e_1, \dots, e_l)$, then, if $\hat{\mathcal{R}}(e)\coloneqq\hat{\mathcal{R}}(\mathbf{x}^i, \mathcal{N}^i, e, f)$, it holds that $\hat{\mathcal{R}}(\bar e) \leq \max(\hat{\mathcal{R}}(e_1), \dots, \hat{\mathcal{R}}(e_l))$.
\end{property}

Two equivalent models predicting the same class for a datapoint will be related to explanations with comparable robustness scores. The data manifold learned by the models influences the robustness score.

\begin{property} \label{p6}
Let $f(\cdot)$ and $g(\cdot)$ be two models with comparable accuracy on the dataset $\mathcal{D}$. If the point $\mathbf{x}^i$ is predicted to belong to the same class by both models, say $\hat{y}_f^i = \hat{y}_g^i$, and let $\hat{\mathcal{R}}(f)\coloneqq \hat{\mathcal{R}}(\mathbf{x}^i, \mathcal{N}^i, e, f)$, then: 
\begin{equation}
    \vert \hat{\mathcal{R}}(f)-\hat{\mathcal{R}}(g)\vert < \delta \text{ with } \delta>0
\end{equation}
\end{property}

We propose the robustness $r$ to be computed as $r(\mathbf{x}^i, \tilde{\mathbf{x}}^i) = \rho(e(\mathbf{x}^i, f), e(\tilde{\mathbf{x}}^i, f))$, where $\rho$ is the Spearman's rho rank correlation coefficient and $e(\mathbf{x}^i, f)$ the explanation of model $f$ prediction of point $\mathbf{x}^i$.
By Property \ref{p1}, it then holds that the robustness $\mathcal{R}$ can be estimated via:
\begin{equation}\label{estimator}
     \hat{\mathcal{R}}(\mathbf{x}^i, \mathcal{N}^i, e, f) = \cfrac{1}{\vert \mathcal{N}^i \vert} \sum \limits_{\tilde{\mathbf{x}}^i \in \mathcal{N}^i} \rho(e(\mathbf{x}^i, f), e(\tilde{\mathbf{x}}^i, f))
\end{equation}
where $\mathcal{N}^i = \{\tilde{\mathbf{x}}^i \vert \tilde{\mathbf{x}}^i = \mathbf{x}^i + \lambda \text{ with } \lambda\in \mathbb{R}^m , \;dist(\mathbf{x}^i,\; \tilde{\mathbf{x}}^i)<\epsilon \, (\epsilon>0) \text{ and } \hat{y}^i = \hat{y}^{\tilde i} \}$. 
By definition, it holds that $0 \leq \hat{\mathcal{R}}(\mathbf{x}^i, \mathcal{N}^i, e, f) \leq 1$.

We will show in Section \ref{experimental-evaluation} that our estimator also satisfies Properties \ref{p2}-\ref{p6}.

\subsection{Neighbourhood Generation}
As remarked in Property \ref{p4}, neighbourhood generation is an influential step in the estimation of the robustness, as the score is averaged over the set $\mathcal{N}^i$. In the previous subsection, we have limited the constraints of the neighbourhood to only consider perturbed points $\tilde{\mathbf{x}}^i$ which are close to the original datapoint $\mathbf{x}^i$ and for which the model's prediction is the same, $\hat{y}^i = \hat{y}^{\tilde i}$. We will consider two possible neighbourhood generation mechanisms which deeply influence the robustness computation. Let us distinguish between numerical and categorical variables, $x_{num}^i$ and $x_{cat}^i$ respectively, as they require different perturbations by construction.

\subsubsection{Random Neighbourhood ($\mathcal{N}_R$)} A first naive approach is the random generation of the neighbourhood, consisting of the addition of random white noise to numerical variables and a random flip of the categorical ones:
\begin{equation}
    \begin{cases}
    \tilde{x}_{num}^i = x_{num}^i + \delta^i \text{ with } \delta^i \leftarrow  \mathcal{N}(0, \sigma^2)\\
    \tilde{x}_{cat}^i = \text{flip}(x_{cat}^i) \text{ with probability } \gamma_{cat}
    \end{cases}
\end{equation}
The flip of a categorical variable entails a random sampling among the possible modalities associated with that variable, the observed value of $x_{cat}^i$ being excluded.

\subsubsection{Medoid-based Neighbourhood ($\mathcal{N}_M$)}
We propose a more refined mechanism that leverages the manifold hypothesis to generate perturbed datapoints which are still on-manifold, as our interest lies in testing non-adversarial perturbations consistent with the observed data distribution. 
Consider a $k$-medoid clustering on $\mathcal{D}_{valid}$, with $k_{medoids}$ selected so that each cluster is, on average, of size $n_k = 10$. For each cluster, the medoid $\mathbf{x}^c$ is used as a representative and its $k_M$ nearest neighbours among the other cluster centres are stored in the set $NN^c = (\mathbf{x}^{1}, \dots, \mathbf{x}^{{k_M}})$. For each point $\mathbf{x}^i\in\mathcal{D}_{test}$ we want to test, the associated cluster $c$ is retrieved. From the corresponding cluster centre neighbours list $NN^c$, one of the medoids is randomly chosen, say $\mathbf{x}^M$. With $\alpha$ and $\alpha_{cat}$ the probabilities of perturbing a numerical and a categorical variable respectively, a perturbation is performed according to the following scheme:
\begin{equation}
    \begin{cases}
    \tilde{x}_{num}^i = (1-\bar \alpha)\cdot x_{num}^i + \bar\alpha \cdot x_{num}^M \text{ with } \bar\alpha \leftarrow Beta(\alpha\cdot100, (1-\alpha)\cdot 100)\\
    \tilde{x}_{cat}^i = \begin{cases}
        x_{cat}^i \text{ with probability } 1-\alpha_{cat} \\
        x_{cat}^M \text{ with probability } \alpha_{cat}
    \end{cases}
    \end{cases}
\end{equation}

With both generating schemes, the resulting neighbourhood should be of at least size $n=100$ to ensure statistical significance. A filtering step is then performed to remove the perturbations for which the model prediction is different from $f(\mathbf{x}^i)$. Hyperparameter tuning is performed on $\theta_1 = (\sigma, \gamma_{cat})$ and $\theta_2 = (\alpha, \alpha_{cat}, k_M)$ to ensure that, on average, at least 95\% of the points are kept. 

\begin{figure}[t] 
\includegraphics[width=\textwidth]{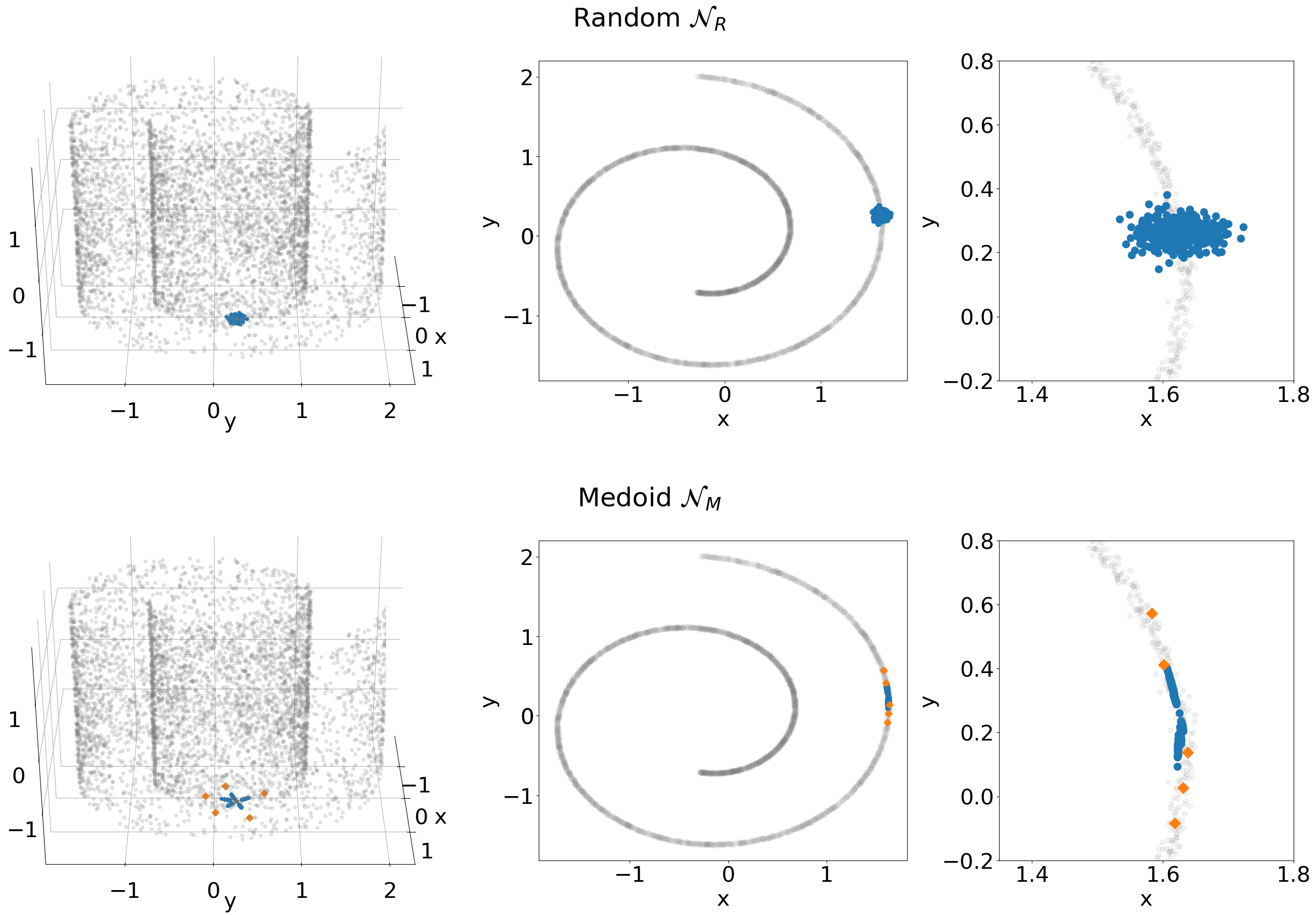}
\caption{An example of neighbourhood generation ($n=500$) on the Swiss roll dataset: the top row depicts the random neighbourhood generation $\mathcal{N}_R$ ($\sigma=0.05$) while the bottom one the medoid-based generation $\mathcal{N}_M$  ($\alpha = 0.3$ and $k_M=5$). Blue datapoints represent the generated perturbations while the orange ones in the bottom row are the $k_M$ neighbouring cluster centres.} \label{swissroll}
\end{figure}

The main differences among the two approaches are presented in Fig. \ref{swissroll}, where the Swiss roll dataset is used as an example. Both schemes were applied to the same datapoint: the left most column represents a 3D visual of the dataset (in the shape of a rolled piece of paper), while the middle one is a view from above. It is easy to note that the random neighbourhood is expanded beyond the Swiss roll spiral shape. This is more evident in the right-most column, where a zoomed-in visualization is proposed: while the random neighbourhood does not follow the data manifold even when a small perturbation is applied ($\sigma = 0.05)$, the medoid-based one remains constantly within bounds despite a larger coefficient being used ($\alpha = 0.3$).

\subsection{Ensemble} 
We present a novel aggregation approach that aims at dealing with the disagreement problem \cite{krishna-lakkaraju} by merging feature importances, focusing on the ranking of the features according to their absolute value. Let $L$ be the number of methods which are being aggregated and let $\mathbf{r}_l^i$ be the $l$-th ranking, a vector of size $m$ storing the indices that would sort the array $\vert \mathbf{a}_l^i \vert$ in decreasing order. Then, let an average attribution $\mathbf{a}_{ens}^i$ be defined by:
\begin{equation}
    a_{ens}^{(i,j)} = \cfrac{\sum_{l=1}^{L} r_l^{(i,j)}\cdot w_l^{(i,j)}}{\sum_{l=1}^{L} w_l^{(i,j)} }\cdot(1+\lambda \bar n^{(i,j)})
\end{equation}
where $w_l$ is the weight corresponding to the $l$-th feature attribution vector, $\lambda = 0.15$ is a penalization term and $\bar n^{(i,j)}$ is the number of methods for which there is a disagreement on the sign of the attribution $a_l^{(i,j)}$ for $l=1, \dots, L$.
The ranking $\mathbf{r}_{ens}^i$ storing the indices that would sort the attribution vector $\mathbf{a}_{ens}^i$ in increasing order is the results of our approach.

This aggregation method is able to deal with practical issues that emerge when considering multiple explanations altogether. First of all, attributions may result in coefficients very close, but not equal, to zero. A zero coefficient implies that the feature is not important towards the prediction, but the lack of a common scale for feature attributions makes it difficult to discriminate between important and unimportant features only based on the absolute value of the corresponding coefficient. We aim at limiting this issue by considering the following weighting scheme where, for the $l$-th methods, $i$-th point and $j$-th feature, it holds:
 \begin{equation}
     w_l^{(i,j)} = \cfrac{\sigma(\mathbf{a}_l^i)}{\sqrt{\vert\overline{\mathbf{a}_l^i}\vert\cdot \vert a_l^{(i,j)}\vert}}
 \end{equation}
with $\mathbf{a}_l^i$ the feature importance vector, $\overline{\mathbf{a}_l^i}$ its average and $\sigma(\mathbf{a}_l^i)$ the standard deviation. 

Smaller values of the feature importance coefficients translate into larger weights, that contribute to moving a feature into the set of non-informative ones of our ranking, effectively capturing the scarcer importance towards the prediction. As coefficients may differ greatly in magnitude between multiple methods, a normalization is then introduced with the consideration of the average and standard deviation within the weighting scheme.

The ensemble is computed as the weighted average of the $L=3$ rankings and a penalization term $\lambda$ is introduced to penalize features that exhibit sign disagreement among the $L$ methods, as we consider it a symptom of feature instability. The penalization term allows us to favour the features for which there is sign concordance ($\bar n^{(i,j)} = 0$), either positive or negative, and when the attribution magnitude is non negligible. Considering how larger weights move features towards the set of non influential ones, the penalization contributes to a lower attribution value for concordant features, effectively capturing their (absolute) relevance and importance towards the prediction.

We will compare our ensemble to another aggregation method in Section \ref{experimental-evaluation}. Inspired by \cite{rieger}, we will simply consider the average of the attributions, when the vectors have norm 1. If $\mathbf{a}_l^i$ is such that $\Vert \mathbf{a}_l^i\Vert_2 = 1$, then 
\begin{equation}
    a_{mean}^{(i,j)} = \cfrac{1}{L}\sum_{l=1}^L a_l^{(i,j)}
\end{equation}
with $\mathbf{a}_{mean}^i$ such that $\Vert \mathbf{a}_{mean}^i \Vert_2 =1$.

One of the disadvantages of using the mean as aggregation is that it may assign a zero attribution even when all $L$ coefficients are non-null. Assume that, for the $j$-th feature, the $L=3$ coefficients derived from the corresponding methods are $(v,v, -2v)$ with $v>0$. The mean will be equal to zero, implying that the feature is non relevant towards the prediction, but the importance for each of the $L$ method highlights a relevance of magnitude at least $v$. Our ensemble is, instead, able to take this into account, penalizing the disagreement but still considering the feature to have some level of relevance towards the prediction.

While we have computed both aggregations with the feature attributions derived from DeepLIFT, Integrated Gradients and LRP (as presented in Subsection \ref{considered_techniques}), both approaches can be easily extended to include a higher number of feature attribution methods.

\subsection{When Can You Trust Your Explanations?}\label{4-4}
Assessing the trustworthiness of explanations on previously unseen datapoints is a non trivial task. While it is possible to compute an estimate of the robustness via Equation \ref{estimator}, we argue that the result may not reflect the true robustness of the considered datapoint, as it may lay in an unstable area of the feature space. We want to verify not only if a datapoint is robust, but also if it lies in a robust area of the feature space: knowing that the neighbourhood is non robust allows us to \textit{doubt} the robustness score of a previously-unseen datapoint. To tackle this issue, we propose the usage of a $k$-nearest neighbours regressor fitted on $\mathcal{D}_{valid}$ robustness scores. For $\mathbf{x}^i\in\mathcal{D}_{test}$, we compute both the robustness score $\hat{\mathcal{R}}(\mathbf{x}^i) \coloneqq \hat{\mathcal{R}}(\mathbf{x}^i, \mathcal{N}^i, e, f)$ via \ref{estimator} and the regressor's prediction $\mathcal{R}_{knn}(\mathbf{x}^i)$. According to a selected threshold value $r_{th}$, it is possible to discriminate between three scenarios:
\begin{enumerate}
    \item if $\hat{\mathcal{R}}(\mathbf{x}^i)\geq r_{th}$ and $\mathcal{R}_{knn}(\mathbf{x}^i)\geq r_{th}$ then $\mathbf{x}^i$ is a robust point;
    \item if $\hat{\mathcal{R}}(\mathbf{x}^i)\geq r_{th}$ and $\mathcal{R}_{knn}(\mathbf{x}^i)<r_{th}$ then $\mathbf{x}^i$ is an uncertain point, as it lies in an uncertain area of the feature space and its robustness should be carefully considered;
    \item if $\hat{\mathcal{R}}(\mathbf{x}^i) < r_{th}$ then $\mathbf{x}^i$ is a non robust point.
\end{enumerate}

The second scenario represents a set of conditions that practitioners should carefully evaluate: despite the robustness score being greater than the selected threshold, the local information derived from the neighbours suggests otherwise. This scenario aims at ringing a bell in the practitioner evaluating a given explanation: knowing that it may be misleading, the analysis will be more careful and require a more detailed human-evaluation of both the prediction and its explanation. 

The usage of a $k$nn regressor instead of a $k$nn classifier allows for independence from the selected robustness threshold $r_{th}$, requiring the model be fitted only once. The number of neighbours $k_R$ is a dataset-dependent hyperparameter and relies on the goodness of approximation of the robustness score by the regressor. 

The selection of the threshold $r_{th}$ is a delicate step of the procedure: as the robustness estimator is bounded by construction in the $[0,1]$ range, it is possible to select a case-specific threshold to discriminate between robust and non robust datapoints. As it will be seen in subsection \ref{hyperparam}, in which hyperparameter selection is discussed in detail, a default value of $r_{th}=0.80$ works well on most of the datasets.

\subsection{A Complete Pipeline}
\begin{figure}[t] 
\includegraphics[width=\textwidth]{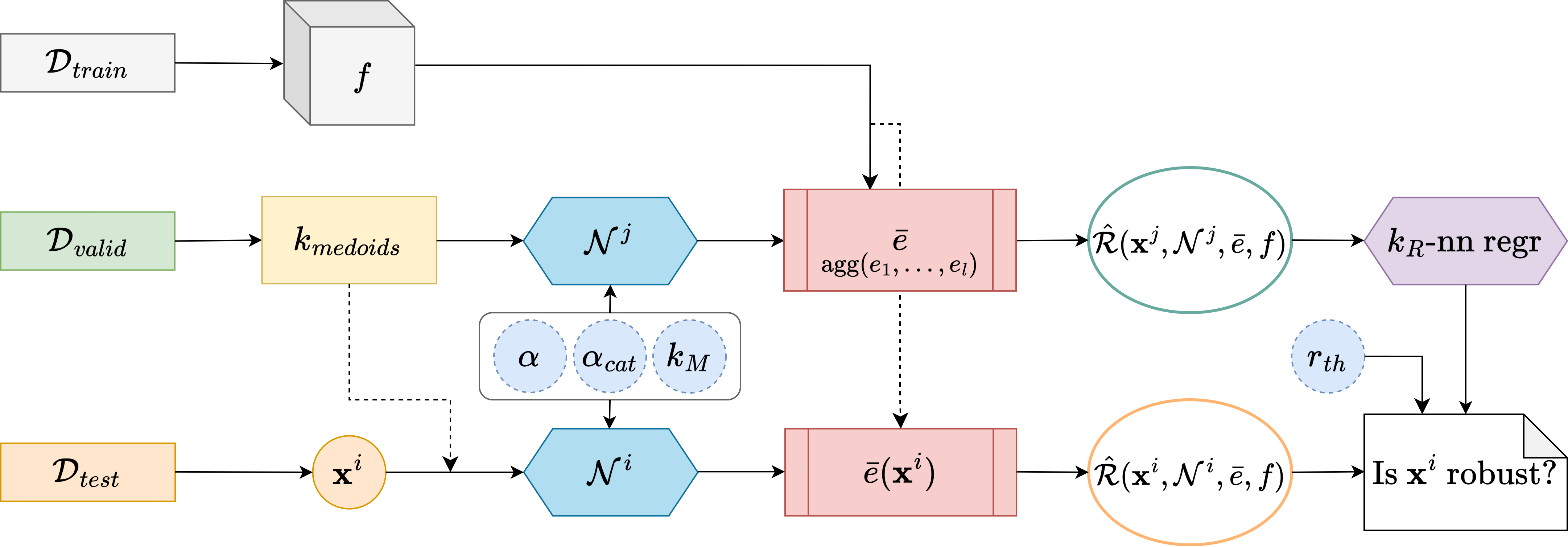}
\caption{The complete pipeline.} \label{pipeline}
\end{figure}

With the considerations presented up to this point, we can discuss the complete framework (Fig. \ref{pipeline}) for robustness evaluation and explanation trustworthiness. 

\begin{enumerate}
    \item Split the dataset into $\mathcal{D}_{train}, \mathcal{D}_{valid}, \mathcal{D}_{test}$ and perform the required preprocessing steps.
    \item Train a neural network on $\mathcal{D}_{train}$.
    \item Perform $k$-medoid clustering on $\mathcal{D}_{valid}$ and compute the $k_M$ nearest neighbours among the medoids.
    \item For each point $\mathbf{x}^j\in \mathcal{D}_{valid}$, generate a neighbourhood $\mathcal{N}^j$ of size $n$.
    \item Compute the attributions of DeepLIFT, Integrated Gradients and LRP and merge them following the ensemble aggregation scheme.
    \item Compute the robustness score via Equation \ref{estimator} for each point $\mathbf{x}^j \in \mathcal{D}_{valid}$.
    \item Use the previously computed robustness scores to train a $k$-nearest neighbours regressor and select an appropriate threshold $r_{th}$.
    \item For each datapoint $\mathbf{x}^i \in \mathcal{D}_{test}$, predict the medoid cluster it belongs to and generate a neighbourhood $\mathcal{N}^i$.
    \item Compute the $L$ feature attribution vectors and the resulting aggregation.
    \item Compute then the robustness score via Equation \ref{estimator} and the predicted robustness via the $k$nn regressor.
    \item Assess the trustworthiness following Subsection \ref{4-4}.
\end{enumerate}

Note that the computational cost of the procedure is dominated by the computation of the individual attributions, as a pass through the network is required for each datapoint, for each perturbation and for each of the three methods. The scaling of the underlying robustness method inherits this complexity.

The advantages of using this framework include the leveraging of an ensemble-based explanation, which captures the signals of $L$ individual explanation methods, and the estimation of the robustness on a carefully constructed neighbourhood. The pipeline is designed to point out possibly non robust points, even when they appear so, ensuring greater trustworthiness in the system and a more aware analysis from a practitioners perspective.

The presented framework is applicable to datasets entirely made of numerical variables, but to correctly consider also categorical ones a small adjustment is required. Categorical features are often preprocessed with a one-hot encoding before a neural network is trained. For example, a feature $x_{cat}$ with four modalities is represented by a vector of the form $(0,0,1,0)$, where the non zero-entry corresponds to the observed modality. A feature attribution method applied to the one-hot encoded feature vector returns a set of importances of the form $(0,0,u,0)$: the non-zero entry is the only one associated with a non-null value. This is to be expected as the attribution methods we are evaluating satisfy the missingness property (Subsection \ref{considered_techniques}). To limit the effects of zero-entries in the Spearman's rho computation, we propose a \textit{reverse encoding} of categorical variables. They are represented by the observed modality and the variable $x_{cat}$ is associated with an attribution $a_l = u$. This allows us to consider feature vectors of size $m$, as in the original feature vector, and to perform a more effective comparison of the robustness. The \textit{reverse encoding} is applied in steps 5 and 9, prior to the ensemble's computation.

\subsection{Validation}\label{validation}
Robustness estimation is often subject to the lack of a ground truth, as the data generation process is not known in real-world applications and comparison against expert-provided explanations can be unfeasible due to the large amount of data being examined. We argue that robust (explanation-wise) points lie in a robust area of the feature space and can be deemed robust even when passed though different models. Our assumption implies that non robust points will exhibit differences both in the explanations and in the predictions of multiple models, as their lack of robustness is a somewhat intrinsic characteristic of the area of the manifold in which they lie. 

Let us consider three neural networks, say $f_1, f_2$ and $f_3$, which have comparable accuracy over $\mathcal{D}_{train}$ and that differ either in the number of hidden layers or neurons per layer. Let $\mathcal{D}_{agree}$ be the subset of datapoints for which models $f_1, f_2, f_3$ predict the same class and $\mathcal{D}_{disagree}$ the subset for which one of the models predicts a different class. Consider a point to be robust if $\hat{\mathcal{R}}(\mathbf{x}^i)\geq r_{th}$ and non-robust otherwise.\footnote[1]{Note that, in this case, we are not considering the classification presented in subsection \ref{4-4}, but only if the robustness score $\hat{\mathcal{R}}$ is above the selected threshold $r_{th}$.}  We propose the validation to follow a ROC/AUC analysis, where the True Positive Rate (TPR) and False Positive Rate (FPR) are defined as:
\begin{equation}
    \text{TPR}=\cfrac{\text{\#\{Robust \& Agree\}}}{\text{\#\{Agree\}}} \;\;\;\;\; \text{FPR}=\cfrac{\text{\#\{Robust \& Disagree\}}}{\text{\#\{Disagree\}}}
\end{equation}

Varying the threshold value $r_{th}$, it is possible to plot the ROC curve and compute the corresponding AUC value.

\section{Experimental Evaluation} \label{experimental-evaluation}
\subsection{Experimental Setting}
We have selected the following publicly-available datasets from the UC Irvine Machine Learning Repository: \texttt{beans}, \texttt{cancer}, \texttt{mushroom}, \texttt{white wine}, \texttt{adult} and \texttt{bank marketing}. We have additionally used the \texttt{heloc} and \texttt{ocean} datasets, following the work of \cite{heloc} and \cite{thor} respectively.
The first four represent toy examples, as the classification tasks are easier to tackle even with non-neural models and present an overall lower number of both features and datapoints. All the datasets propose binary or multiclass classification tasks and contain both numerical and categorical variables. 

We have relied on the Python libraries \texttt{pytorch} and \texttt{captum} for the implementation of our approach. The former was used for the training and usage of the nets while the latter, a \texttt{pytorch}-compatible explainability framework developed by Meta, was applied for the retrieval of the attribution vectors. The full implementation is available on Github.\footnote[2]{\url{https://github.com/ilariavascotto/XAI_robustness_analysis}}

For all datasets, the following preprocessing steps were performed:
\begin{itemize}
    \item Standardization of numerical variables and one-hot encoding of  the categorical ones.
    \item Removal of correlated features. Spearman's rank correlation coefficient was used for numerical variables while the normalized mutual information criterion for the categorical ones. The removal of highly correlated features ensures that neighbourhood generation is more aligned with the data distribution and that the computed attributions are non distorted by unconsidered correlations among the features. 
    \item Softmax in the final layer of the net, also for binary classification examples. It ensures stability during attribution computations, as higher relevances are backpropagated. This also reduces the effects of the vanishing gradient problem, when the attribution is returned as a zero-vector, as the signal is lost when the output score is propagated through the net.
    \item Selection of the \texttt{gamma} rule for LRP attributions. It was chosen as it was the rule minimizing the vanishing gradient problem, and it was applied to all the layers of the nets (as they are all linear layers). 
\end{itemize}

As anticipated in Subsection \ref{validation}, we trained three neural networks per dataset, say Model 1, 2 and 3, with comparable accuracy scores (Appendix \ref{appendix}, Table \ref{dataset}). Model 1 represents the baseline model, model 2 has more layers and more neurons in each layer while model 3 is a more compact version, with fewer layers and neurons. The ReLU activation function was used in all cases, with the exception of the \texttt{ocean} dataset, where tanh was used, following the structure in \cite{thor}.

\subsubsection{Hyperparameter Selection} \label{hyperparam} Default values for the neighbourhood generation hyperparameters are $\theta_1 = (\sigma= 0.05, \gamma_{cat} = 0.05)$ for the random generation $\mathcal{N}_R$ and $\theta_2 = (\alpha = 0.05, \alpha_{cat} = 0.05, k_M = 5)$ for the medoid-based one $\mathcal{N}_M$. In both cases, the neighbourhood should be of size at least $n=100$ (we have set $n=100$ in our experiments) and dataset-specific hyperparameters are set via a grid search, ensuring that at least 95\% of the generated datapoints are kept within the neighbourhood, ie. they are predicted to belong to the same class as the original datapoint. 

The number of neighbours $k_R$ to be used in the $k$nn regressor is chosen as the one minimizing the approximation error over the robustness scores derived from all three nets in each dataset. A good default value is $k_R = 7$.

The robustness threshold $r_{th}$ is selected by looking at the distribution of robust, non robust and uncertain datapoints at varying levels of the threshold. In particular, it is chosen as the threshold value that corresponds to the first inflection point of the robust percentage curve. The default value $r_{th} = 0.80$ works well in most scenarios.

Dataset-specific hyperparameters can be found in Appendix \ref{appendix}, Table \ref{hyperparameters}.

\subsection{Results}
We will begin the discussion focusing on the results derived from Model 1, that acts as our baseline, on the test set $\mathcal{D}_{test}$.

\begin{figure}[t] 
\includegraphics[width=\textwidth]{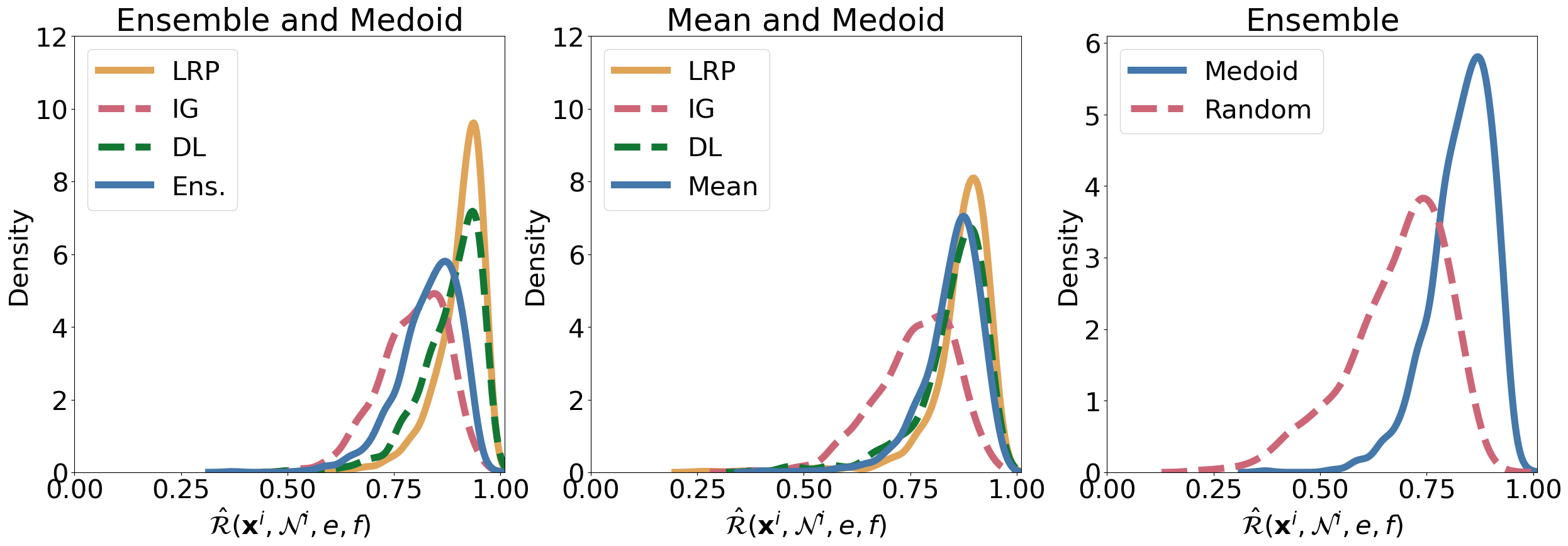}
\caption{The robustness score distribution for \texttt{adult} dataset of the ensemble (left) and of the mean (centre) with medoid-based neighbourhood generation; random and medoid-based neighbourhood generations are compared for the ensemble aggregation (right).} \label{robustness}
\end{figure}

Fig. \ref{robustness} depicts the robustness score distribution derived from the ensemble (left) and the mean (centre) aggregations with the medoid-based neighbourhood generation mechanism on the \texttt{adult} dataset. They are compared with the robustness scores derived from the individual XAI approaches, taking into account the nature of the aggregation. In particular, the ensemble is compared to the robustness computed on the feature importance vectors in absolute value, to mimic the reasoning of the ensemble's construction, while the mean is compared to the feature vectors with sign. As stated by Property \ref{p5}, in both cases the aggregated explanation robustness acts as an average of the robustness of the individual approaches and is limited by their span. In this example, Integrated Gradients is on average the least robust method while DeepLIFT and LRP present grater values of the estimated robustness. Note that this behaviour is dataset dependent: Integrated Gradients is not, in general, the least robust method. The aggregated explanations, either with the ensemble or the mean, represent an advantage over the use of individual approaches. In particular, the aggregation acts as a conservative explanation, that takes into account the individual method's robustness and their feature-wise agreement. While the average robustness may be lower than that of some of the considered approaches, it is able to take into account the possible undesirable effects of a less robust and disagreeing method, flagging possible untrustworthiness for a given datapoint. 

The rightmost part of Fig. \ref{robustness} depicts the comparison between the effects of two neighbourhood generating schemes on the ensemble non-adversarial robustness. As previously shown in Fig. \ref{swissroll}, the random neighbourhood consists of datapoints which are off-manifold, while the medoid-based one is constructed to remain on-manifold, mimicking the effects of non-adversarial perturbations. This reflects into larger average robustness over the test set of the latter neighbourhood over the random one, validating that the robustness estimator with the ensemble consistently satisfies Property \ref{p4}. The same property is satisfied also by the mean aggregation and the individual XAI methods when tested individually.

\begin{table}[t]
\caption{When can you trust your explanations? Comparison of the robustness scores with ensemble and mean aggregations with $r_{th}=0.80$.}
\label{table1}
\centering
\begin{tabular}{cc|ccc|ccc|}
\cline{3-8}
 &  & \multicolumn{3}{c|}{Ensemble} & \multicolumn{3}{c|}{Mean} \\ \hline
\multicolumn{1}{|c|}{Dataset} & $\mathcal{D}_{test}$ & \multicolumn{1}{c|}{Robust} & \multicolumn{1}{c|}{Uncertain} & Non Robust & \multicolumn{1}{c|}{Robust} & \multicolumn{1}{c|}{Uncertain} & Non Robust \\ \hline
\multicolumn{1}{|c|}{beans} & 500 & \multicolumn{1}{c|}{74.6\%} & \multicolumn{1}{c|}{7.2\%} & 18.2\% & \multicolumn{1}{c|}{93.0\%} & \multicolumn{1}{c|}{2.2\%} & 4.8\% \\ \hline
\multicolumn{1}{|c|}{cancer} & 50 & \multicolumn{1}{c|}{62.0\%} & \multicolumn{1}{c|}{12.0\%} & 26.0\% & \multicolumn{1}{c|}{100.0\%} & \multicolumn{1}{c|}{0.0\%} & 0.0\% \\ \hline
\multicolumn{1}{|c|}{mushroom} & 400 & \multicolumn{1}{c|}{0.0\%} & \multicolumn{1}{c|}{1.2\%} & 98.8\% & \multicolumn{1}{c|}{0.0\%} & \multicolumn{1}{c|}{0.0\%} & 100.0\% \\ \hline
\multicolumn{1}{|c|}{white wine} & 200 & \multicolumn{1}{c|}{7.5\%} & \multicolumn{1}{c|}{26.5\%} & 66.0\% & \multicolumn{1}{c|}{56.5\%} & \multicolumn{1}{c|}{11.0\%} & 32.5\% \\ \hline
\multicolumn{1}{|c|}{adult} & 1000 & \multicolumn{1}{c|}{63.4\%} & \multicolumn{1}{c|}{7.1\%} & 29.5\% & \multicolumn{1}{c|}{72.6\%} & \multicolumn{1}{c|}{7.3\%} & 20.1\% \\ \hline
\multicolumn{1}{|c|}{bank} & 1000 & \multicolumn{1}{c|}{65.2\%} & \multicolumn{1}{c|}{7.1\%} & 27.7\% & \multicolumn{1}{c|}{44.6\%} & \multicolumn{1}{c|}{16.4\%} & 39.0\% \\ \hline
\multicolumn{1}{|c|}{heloc} & 500 & \multicolumn{1}{c|}{62.8\%} & \multicolumn{1}{c|}{9.2\%} & 28.0\% & \multicolumn{1}{c|}{60.6\%} & \multicolumn{1}{c|}{14.0\%} & 25.4\% \\ \hline
\multicolumn{1}{|c|}{ocean} & 10000 & \multicolumn{1}{c|}{79.1\%} & \multicolumn{1}{c|}{5.6\%} & 15.3\% & \multicolumn{1}{c|}{53.5\%} & \multicolumn{1}{c|}{16.0\%} & 30.5\% \\ \hline
\end{tabular}
\end{table}

Table \ref{table1} presents the classification of the test set into robust, uncertain and non robust datapoints (as per Subsection \ref{4-4}). For comparability, we set $r_{th} = 0.80$ for all datasets and both aggregation methods, even if the ensemble often requires lower values of $r_{th}$ compared to the mean (see Appendix \ref{appendix}). Medoid-based neighbourhoods were used, as it was shown in Fig. \ref{robustness} that they produce larger robustness scores. 

For almost all the datasets, robust datapoints are the majority, with the exception of the \texttt{white wine} dataset with the ensemble and \texttt{mushroom} dataset with both aggregations. This is due to the selection of the threshold value equal for all datasets, as carefully selected dataset-specific thresholds would be lower than $0.80$.  The percentages of uncertain and non robust datapoints are non-negligible. In particular, we consider uncertain datapoints to be the most interesting ones to investigate. They represent areas of the feature space where the robustness estimation is uncertain (as per Property \ref{p3}) and should be carefully considered during a practical evaluation.

\begin{figure}[t] 
\centering
\includegraphics[width=0.85\textwidth]{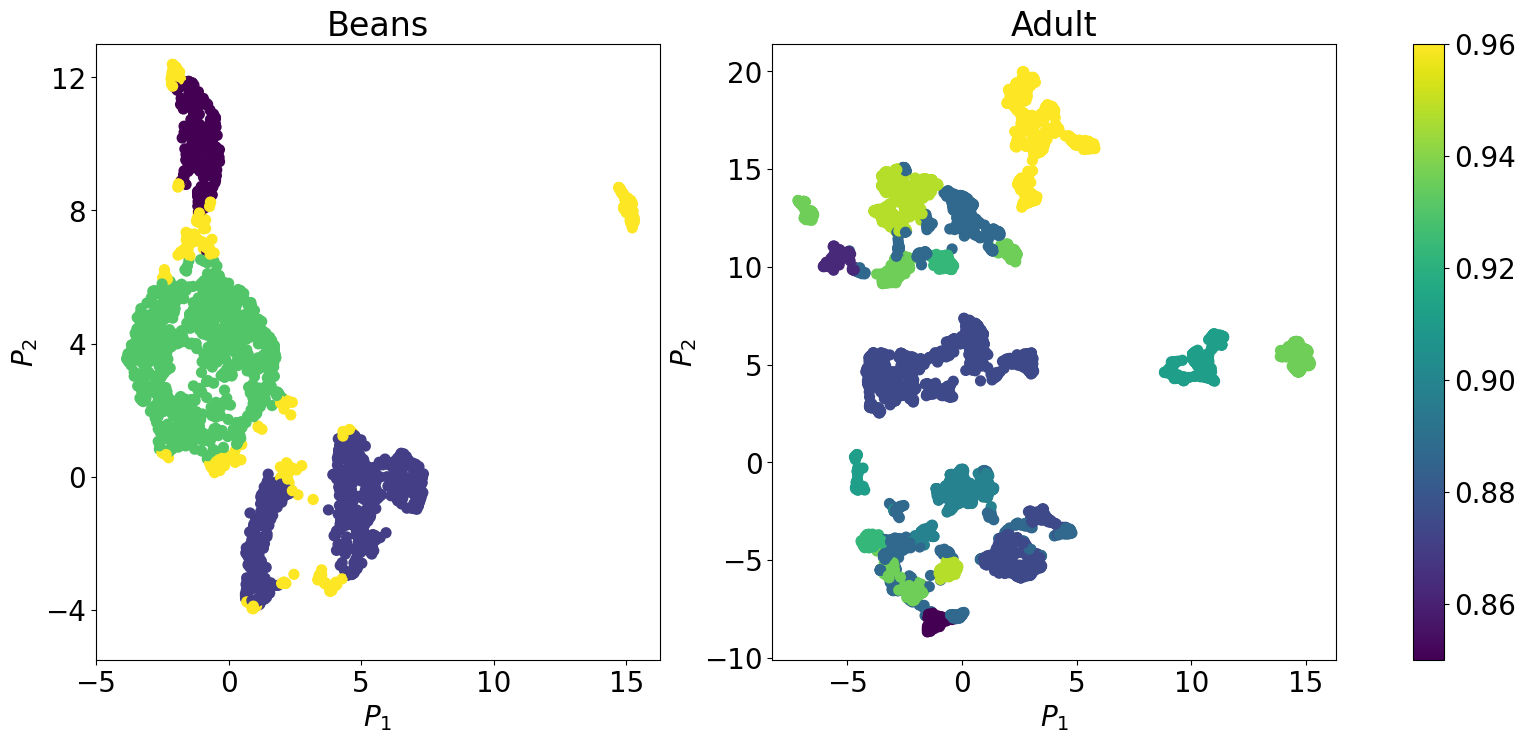}
\caption{HDBSCAN clusters, coloured by mean robustness scores, over the two dimensional UMAP projections ($P_1, P_2$): \texttt{beans} dataset (left) and \texttt{adult} dataset (right).} \label{umap}
\end{figure}

In Fig. \ref{umap} we present a two dimensional visualization of the UMAP \cite{umap} projections of the validation set $\mathcal{D}_{valid}$, clustered with HDBSCAN algorithm \cite{hdbscan}, where each cluster is coloured by its mean robustness score. UMAP is a dimensionality reduction technique that allows us to visualize a lower dimensional projection of the data. We applied the density-based clustering algorithm HDBSCAN to such projections, producing clusters without the need of setting hyperparameters for the desired number of clusters, as in $k$-means for example. As can be seen in the figure, the projected data space may be more or less complex according to the dataset being analysed. Robustness homogeneity within the derived clusters support the claims of Property \ref{p2}, in which close points are expected to exhibit smaller differences between their robustness scores. This allows us to support the use of the $k$nn regressor to estimate local robustness scores as points are naturally grouped in clusters with similar robustness scores. 

\begin{table*}[t] 
\caption{Comparison of ensemble's robustness classification (with $r_{th}=0.80$) according to Model 1,2,3 concordance in the prediction.}
\label{table2}
\centering
\begin{tabular}{c|ccc|ccc|}
\cline{2-7}
 & \multicolumn{3}{c|}{Robust} & \multicolumn{3}{c|}{Non Robust} \\ \hline
\multicolumn{1}{|c|}{Dataset} & \multicolumn{1}{c|}{Agree} & \multicolumn{1}{c|}{Disagree} & N. points & \multicolumn{1}{c|}{Agree} & \multicolumn{1}{c|}{Disagree} & N. points \\ \hline
\multicolumn{1}{|c|}{beans} & \multicolumn{1}{c|}{95.35\%} & \multicolumn{1}{c|}{4.65\%} & 409 & \multicolumn{1}{c|}{95.60\%} & \multicolumn{1}{c|}{4.40\%} & 91 \\ \hline
\multicolumn{1}{|c|}{cancer} & \multicolumn{1}{c|}{91.89\%} & \multicolumn{1}{c|}{8.11\%} & 37 & \multicolumn{1}{c|}{92.31\%} & \multicolumn{1}{c|}{7.69\%} & 13 \\ \hline
\multicolumn{1}{|c|}{mushroom} & \multicolumn{1}{c|}{100.00\%} & \multicolumn{1}{c|}{0.00\%} & 5 & \multicolumn{1}{c|}{100.00\%} & \multicolumn{1}{c|}{0.00\%} & 395 \\ \hline
\multicolumn{1}{|c|}{white wine} & \multicolumn{1}{c|}{94.11\%} & \multicolumn{1}{c|}{5.88\%} & 68 & \multicolumn{1}{c|}{92.42\%} & \multicolumn{1}{c|}{7.58\%} & 132 \\ \hline
\multicolumn{1}{|c|}{adult} & \multicolumn{1}{c|}{93.76\%} & \multicolumn{1}{c|}{6.24\%} & 705 & \multicolumn{1}{c|}{80.68\%} & \multicolumn{1}{c|}{19.32\%} & 295 \\ \hline
\multicolumn{1}{|c|}{bank} & \multicolumn{1}{c|}{95.99\%} & \multicolumn{1}{c|}{4.01\%} & 723 & \multicolumn{1}{c|}{90.25\%} & \multicolumn{1}{c|}{9.75\%} & 277\\ \hline
\multicolumn{1}{|c|}{heloc} & \multicolumn{1}{c|}{82.22\%} & \multicolumn{1}{c|}{17.78\%} & 360 & \multicolumn{1}{c|}{56.43\%} & \multicolumn{1}{c|}{43.57\%} & 140 \\ \hline
\multicolumn{1}{|c|}{ocean} & \multicolumn{1}{c|}{85.21\%} & \multicolumn{1}{c|}{14.8\%} & 8475 & \multicolumn{1}{c|}{80.33\%} & \multicolumn{1}{c|}{19.67\%} & 1525 \\ \hline
\end{tabular}
\end{table*}

As introduced in Subsection \ref{validation}, the validation of robustness estimations is subject to the lack of a ground truth. Table \ref{table2} shows how the percentage of points over which the three models (Model 1, 2 and 3) (dis)agree varies according to the predicted robustness of the datapoints, with $r_{th} = 0.80$ for all datasets. The ensemble robustness is computed with the medoid-based neighbourhood. Note that the \texttt{mushroom} dataset is a particular case, as all three methods reach 100\% accuracy and are, therefore, always agreeing. It can be seen that the percentage of disagreeing points within the non-robust ones is, for most datasets, greater than that of the robust ones. This supports our validation proposal, as we consider the disagreement in the predictions to be a symptom of non robustness, as per Property \ref{p6}.

\begin{figure}[t] 
\includegraphics[width=\textwidth]{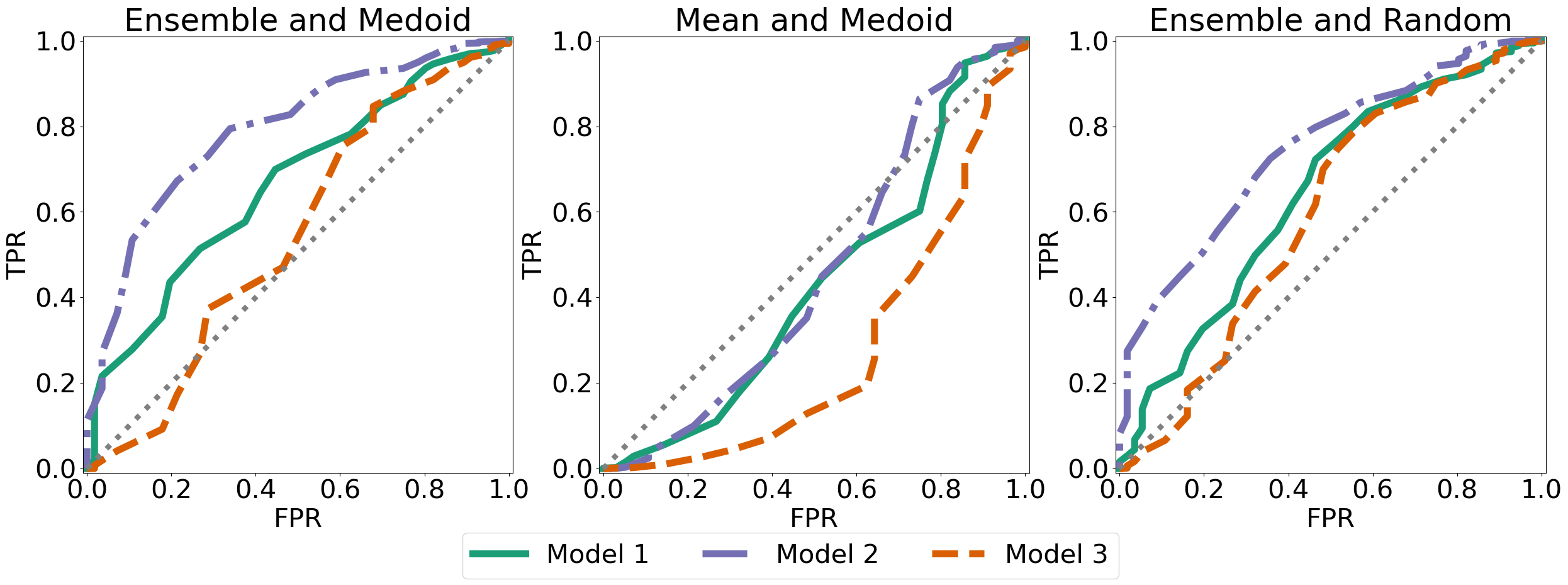}
\caption{ROC curves on the \texttt{bank} dataset with ensemble aggregation and medoid-based neighbourhood (left), mean aggregation and medoid-based neighbourhood (centre) and ensemble aggregation with random neighbourhood (right). The gray dotted line represents the bisecting line.} \label{roc}
\end{figure}

The ROC/AUC analysis presented in Subsection \ref{validation} allows us to jointly consider the aggregation method and the neighbourhood generation scheme which better fit the dataset at hand. Fig. \ref{roc} depicts the ROC curves of the three models in varying scenarios: Model 2 is consistently associated with the highest ROC curve, suggesting that, for the \texttt{bank} dataset, a deeper net is better able to propose robust explanations. Moreover, the ROC curves are preferable for all three models with the ensemble aggregation and the medoid-based neighbourhood, while the mean aggregation (as shown in the central plot) presents ROC curves even below the bisecting line.

This is further confirmed by the results presented in Table \ref{auc}, where the average AUC is computed for the four possible combinations of aggregation method and neighbourhood generation. The dataset-wise maximum values of AUC are presented in bold, showing how the ensemble is, on average, preferable with respect to the mean aggregation and how the medoid-based neighbourhood is related to larger AUC values in most examples. 
The \texttt{mushroom} dataset represents, as in Table \ref{table2}, a peculiar case: having only agreeing datapoints the ROC and AUC cannot be computed and are therefore exempt from this analysis.

\begin{table*}[t]
\caption{Average AUC value of Model 1, 2, 3 for both aggregation types and neighbourhoods. For each dataset, the largest average AUC is presented in bold.}
\label{auc}
\centering
\begin{tabular}{c|cc|cc|}
\cline{2-5}
 & \multicolumn{2}{c|}{Medoid} & \multicolumn{2}{c|}{Random} \\ \hline
\multicolumn{1}{|c|}{Dataset} & \multicolumn{1}{c|}{Ensemble} & Mean & \multicolumn{1}{c|}{Ensemble} & Mean \\ \hline
\multicolumn{1}{|c|}{beans} & \multicolumn{1}{c|}{\textbf{0.5711}} & 0.4854 & \multicolumn{1}{c|}{0.4842} & 0.4067 \\ \hline
\multicolumn{1}{|c|}{cancer} & \multicolumn{1}{c|}{0.7346} & \textbf{0.7400} & \multicolumn{1}{c|}{0.6766} & 0.6612 \\ \hline
\multicolumn{1}{|c|}{mushroom} & \multicolumn{1}{c|}{0.0000} & 0.0000 & \multicolumn{1}{c|}{0.0000} & 0.0000 \\ \hline
\multicolumn{1}{|c|}{white wine} & \multicolumn{1}{c|}{0.4762} & 0.5931 & \multicolumn{1}{c|}{0.4819} & \textbf{0.6614} \\ \hline
\multicolumn{1}{|c|}{adult} & \multicolumn{1}{c|}{0.7018} & 0.6695 & \multicolumn{1}{c|}{\textbf{0.8284}} & 0.8077 \\ \hline
\multicolumn{1}{|c|}{bank} & \multicolumn{1}{c|}{\textbf{0.6670}} & 0.3883 & \multicolumn{1}{c|}{0.6612} & 0.4696 \\ \hline
\multicolumn{1}{|c|}{heloc} & \multicolumn{1}{c|}{0.6640} & \textbf{0.6673} & \multicolumn{1}{c|}{0.6262} & 0.5875 \\ \hline
\multicolumn{1}{|c|}{ocean} & \multicolumn{1}{c|}{\textbf{0.5194}} & 0.5128 & \multicolumn{1}{c|}{0.4350} & 0.4053 \\ \hline
\end{tabular}
\end{table*}

\section{Conclusions and Future Work} \label{conclusion}
We presented a novel framework to test explanation robustness, introducing a new neighbourhood generation mechanism, an ensemble approach to merging explanations and a validation test. We have shown that our robustness estimator satisfies the \textit{desiderata} \ref{p1}-\ref{p6} and it overcomes the limitations of other metrics (Section \ref{related-work}). In practical applications, our approach would aid practitioners in understanding the quality of an explanation in terms of its robustness, allowing questioning on the proposed results when the framework flags a datapoint as \textit{uncertain}. 

We have proposed our work targeting neural networks, but the approach is agnostic in nature with respect to both the investigated model and the XAI techniques being applied. Future steps include a first generalization of the proposal on different classes of machine learning models, such as tree-based ones, assuming that local feature importance approaches are available for testing. Future work will also aim at better investigating the relationship between robustness and adversarial attacks. In particular, we aim at assessing the defence ability of our ensemble - as aggregated explanations have proved to be more resilient to adversarial attacks in numerous contexts - and to validate whether our robustness estimator is able to detect attacks. Lastly, we wish to investigate how our explanations could be used to increase the robustness of classifiers, as in \cite{rasouli}.

\begin{credits}
\subsubsection{\ackname} 
This study was partially carried out within the PNRR research activities of the consortium iNEST funded by the European Union Next-GenerationEU (PNRR, Missione 4 Componente 2, Investimento 1.5– D.D. 1058 23/06/2022, ECS 00000043).

\subsubsection{\discintname}
The authors have no competing interests to declare that are
relevant to the content of this article. 
\end{credits}

\appendix

\section{Dataset Description and Hyperparameters} \label{appendix}

\begin{table}[h]
\renewcommand\thetable{A} 
\caption{Dataset description and model accuracy.}
\centering

\label{dataset}
\begin{tabular}{c|ccccccc|ccc|}
\cline{2-11}
 & \multicolumn{7}{c|}{Dataset details} & \multicolumn{3}{c|}{Accuracy (\%) - $\mathcal{D}_{train}$} \\ \hline
\multicolumn{1}{|c|}{Dataset} & \multicolumn{1}{c|}{Classes} & \multicolumn{1}{c|}{\#Num} & \multicolumn{1}{c|}{\#Cat} & \multicolumn{1}{c|}{$\mathcal{D}_{train}$} & \multicolumn{1}{c|}{$\mathcal{D}_{valid}$} & \multicolumn{1}{c|}{$\mathcal{D}_{test}$} & $k_{medoids}$ & \multicolumn{1}{c|}{Model 1} & \multicolumn{1}{c|}{Model 2} & Model 3 \\ \hline
\multicolumn{1}{|c|}{beans} & \multicolumn{1}{c|}{7} & \multicolumn{1}{c|}{7} & \multicolumn{1}{c|}{0} & \multicolumn{1}{c|}{10888} & \multicolumn{1}{c|}{2223} & \multicolumn{1}{c|}{500} & 225 & \multicolumn{1}{c|}{93.41} & \multicolumn{1}{c|}{93.44} & 89.47 \\ \hline
\multicolumn{1}{|c|}{cancer} & \multicolumn{1}{c|}{2} & \multicolumn{1}{c|}{15} & \multicolumn{1}{c|}{0} & \multicolumn{1}{c|}{397} & \multicolumn{1}{c|}{121} & \multicolumn{1}{c|}{50} & 10 & \multicolumn{1}{c|}{99.5} & \multicolumn{1}{c|}{99.75} & 99.24 \\ \hline
\multicolumn{1}{|c|}{mushroom} & \multicolumn{1}{c|}{2} & \multicolumn{1}{c|}{0} & \multicolumn{1}{c|}{21} & \multicolumn{1}{c|}{6498} & \multicolumn{1}{c|}{1225} & \multicolumn{1}{c|}{400} & 120 & \multicolumn{1}{c|}{100.00} & \multicolumn{1}{c|}{100.00} & 100.00 \\ \hline
\multicolumn{1}{|c|}{white wine} & \multicolumn{1}{c|}{2} & \multicolumn{1}{c|}{9} & \multicolumn{1}{c|}{0} & \multicolumn{1}{c|}{3918} & \multicolumn{1}{c|}{780} & \multicolumn{1}{c|}{200} & 80 & \multicolumn{1}{c|}{89.23} & \multicolumn{1}{c|}{89.10} & 86.55 \\ \hline
\multicolumn{1}{|c|}{adult} & \multicolumn{1}{c|}{2} & \multicolumn{1}{c|}{5} & \multicolumn{1}{c|}{7} & \multicolumn{1}{c|}{36177} & \multicolumn{1}{c|}{8045} & \multicolumn{1}{c|}{1000} & 1000 & \multicolumn{1}{c|}{91.39} & \multicolumn{1}{c|}{91.38} & 91.09 \\ \hline
\multicolumn{1}{|c|}{bank} & \multicolumn{1}{c|}{2} & \multicolumn{1}{c|}{5} & \multicolumn{1}{c|}{9} & \multicolumn{1}{c|}{36168} & \multicolumn{1}{c|}{8043} & \multicolumn{1}{c|}{1000} & 1000 & \multicolumn{1}{c|}{91.99} & \multicolumn{1}{c|}{91.76} & 91.45 \\ \hline
\multicolumn{1}{|c|}{heloc} & \multicolumn{1}{c|}{2} & \multicolumn{1}{c|}{14} & \multicolumn{1}{c|}{2} & \multicolumn{1}{c|}{8367} & \multicolumn{1}{c|}{1592} & \multicolumn{1}{c|}{500} & 130 & \multicolumn{1}{c|}{85.50} & \multicolumn{1}{c|}{85.57} & 85.01 \\ \hline
\multicolumn{1}{|c|}{ocean} & \multicolumn{1}{c|}{6} & \multicolumn{1}{c|}{8} & \multicolumn{1}{c|}{0} & \multicolumn{1}{c|}{109259} & \multicolumn{1}{c|}{30328} & \multicolumn{1}{c|}{10000} & 3000 & \multicolumn{1}{c|}{92.04} & \multicolumn{1}{c|}{87.88} & 92.27 \\ \hline
\end{tabular}
\end{table}

\begin{table}[h]
\renewcommand\thetable{B} 
\caption{Dataset-specific hyperparameter selection.}
\centering
\label{hyperparameters}
\begin{tabular}{c|ccccccc|l|cccccc|}
\cline{2-15}
\textbf{} & \multicolumn{7}{c|}{\textbf{$\mathcal{N}_M$}} & \textbf{} & \multicolumn{6}{c|}{\textbf{$\mathcal{N}_R$}} \\ \cline{2-15} 
\textbf{} & \multicolumn{3}{c|}{\textbf{Neighbourhood}} & \multicolumn{2}{c|}{\textbf{Ensemble}} & \multicolumn{2}{c|}{\textbf{Mean}} & \textbf{} & \multicolumn{2}{c|}{\textbf{Neighbourhood}} & \multicolumn{2}{c|}{\textbf{Ensemble}} & \multicolumn{2}{c|}{\textbf{Mean}} \\ \hline
\multicolumn{1}{|c|}{Dataset} & \multicolumn{1}{c|}{\textbf{$\alpha$}} & \multicolumn{1}{c|}{\textbf{$\alpha_{cat}$}} & \multicolumn{1}{c|}{\textbf{$k_M$}} & \multicolumn{1}{c|}{\textbf{$k_R$}} & \multicolumn{1}{c|}{\textbf{$r_{th}$}} & \multicolumn{1}{c|}{\textbf{$k_R$}} & \textbf{$r_{th}$} & \textbf{} & \multicolumn{1}{c|}{\textbf{$\sigma$}} & \multicolumn{1}{c|}{\textbf{$\gamma_{cat}$}} & \multicolumn{1}{c|}{\textbf{$k_R$}} & \multicolumn{1}{c|}{\textbf{$r_{th}$}} & \multicolumn{1}{c|}{\textbf{$k_R$}} & \textbf{$r_{th}$} \\ \hline
\multicolumn{1}{|c|}{beans} & \multicolumn{1}{c|}{0.10} & \multicolumn{1}{c|}{-} & \multicolumn{1}{c|}{10} & \multicolumn{1}{c|}{9} & \multicolumn{1}{c|}{0.85} & \multicolumn{1}{c|}{9} & 0.90 &  & \multicolumn{1}{c|}{0.02} & \multicolumn{1}{c|}{-} & \multicolumn{1}{c|}{5} & \multicolumn{1}{c|}{0.75} & \multicolumn{1}{c|}{7} & 0.80 \\ \hline
\multicolumn{1}{|c|}{cancer} & \multicolumn{1}{c|}{0.10} & \multicolumn{1}{c|}{-} & \multicolumn{1}{c|}{4} & \multicolumn{1}{c|}{11} & \multicolumn{1}{c|}{0.85} & \multicolumn{1}{c|}{9} & 0.90 &  & \multicolumn{1}{c|}{0.10} & \multicolumn{1}{c|}{-} & \multicolumn{1}{c|}{5} & \multicolumn{1}{c|}{0.55} & \multicolumn{1}{c|}{7} & 0.65 \\ \hline
\multicolumn{1}{|c|}{mushroom} & \multicolumn{1}{c|}{-} & \multicolumn{1}{c|}{0.15} & \multicolumn{1}{c|}{10} & \multicolumn{1}{c|}{7} & \multicolumn{1}{c|}{0.70} & \multicolumn{1}{c|}{11} & 0.60 &  & \multicolumn{1}{c|}{-} & \multicolumn{1}{c|}{0.15} & \multicolumn{1}{c|}{9} & \multicolumn{1}{c|}{0.70} & \multicolumn{1}{c|}{11} & 0.60 \\ \hline
\multicolumn{1}{|c|}{white wine} & \multicolumn{1}{c|}{0.15} & \multicolumn{1}{c|}{-} & \multicolumn{1}{c|}{5} & \multicolumn{1}{c|}{11} & \multicolumn{1}{c|}{0.85} & \multicolumn{1}{c|}{7} & 0.80 &  & \multicolumn{1}{c|}{0.03} & \multicolumn{1}{c|}{-} & \multicolumn{1}{c|}{9} & \multicolumn{1}{c|}{0.45} & \multicolumn{1}{c|}{7} & 0.65 \\ \hline
\multicolumn{1}{|c|}{adult} & \multicolumn{1}{c|}{0.05} & \multicolumn{1}{c|}{0.05} & \multicolumn{1}{c|}{5} & \multicolumn{1}{c|}{9} & \multicolumn{1}{c|}{0.80} & \multicolumn{1}{c|}{9} & 0.80 &  & \multicolumn{1}{c|}{0.05} & \multicolumn{1}{c|}{0.05} & \multicolumn{1}{c|}{5} & \multicolumn{1}{c|}{0.70} & \multicolumn{1}{c|}{5} & 0.70 \\ \hline
\multicolumn{1}{|c|}{bank} & \multicolumn{1}{c|}{0.05} & \multicolumn{1}{c|}{0.10} & \multicolumn{1}{c|}{5} & \multicolumn{1}{c|}{7} & \multicolumn{1}{c|}{0.80} & \multicolumn{1}{c|}{11} & 0.80 &  & \multicolumn{1}{c|}{0.05} & \multicolumn{1}{c|}{0.10} & \multicolumn{1}{c|}{7} & \multicolumn{1}{c|}{0.75} & \multicolumn{1}{c|}{9} & 0.75 \\ \hline
\multicolumn{1}{|c|}{heloc} & \multicolumn{1}{c|}{0.05} & \multicolumn{1}{c|}{0.05} & \multicolumn{1}{c|}{5} & \multicolumn{1}{c|}{15} & \multicolumn{1}{c|}{0.80} & \multicolumn{1}{c|}{13} & 0.80 &  & \multicolumn{1}{c|}{0.03} & \multicolumn{1}{c|}{0.10} & \multicolumn{1}{c|}{11} & \multicolumn{1}{c|}{0.45} & \multicolumn{1}{c|}{5} & 0.60 \\ \hline
\multicolumn{1}{|c|}{ocean} & \multicolumn{1}{c|}{0.05} & \multicolumn{1}{c|}{-} & \multicolumn{1}{c|}{5} & \multicolumn{1}{c|}{5} & \multicolumn{1}{c|}{0.65} & \multicolumn{1}{c|}{5} & 0.75 &  & \multicolumn{1}{c|}{0.001} & \multicolumn{1}{c|}{-} & \multicolumn{1}{c|}{5} & \multicolumn{1}{c|}{0.75} & \multicolumn{1}{c|}{5} & 0.65 \\ \hline
\end{tabular}
\end{table}

%
%
%
\bibliographystyle{splncs04}
\bibliography{bibliography}

\end{document}